\documentclass[10pt,times,mathptm,conference]{IEEEtran}

\usepackage[pdftex]{graphicx}
\DeclareGraphicsExtensions{.pdf,.jpeg,.png}

\usepackage{hyperref}       
\usepackage{url}            
\usepackage{booktabs}       
\usepackage{array}          
\usepackage{nicefrac}       
\usepackage{microtype}      
\usepackage{wrapfig}        
\usepackage{subcaption}
\usepackage{adjustbox}
\usepackage{multirow}

\newcolumntype{C}{>{\centering\arraybackslash}m{\dimexpr.1\linewidth-1\tabcolsep}}
\newcolumntype{M}{>{\centering\arraybackslash}m{\dimexpr.225\linewidth-1\tabcolsep}}

\begin{document}
%
\title{LinkNet: Exploiting Encoder Representations for Efficient Semantic Segmentation}


\author{\IEEEauthorblockN{Abhishek Chaurasia}
\IEEEauthorblockA{School of Electrical and Computer Engineering\\
Purdue University\\
West Lafayette, USA\\
Email: aabhish@purdue.edu}
\and
\IEEEauthorblockN{Eugenio Culurciello}
\IEEEauthorblockA{Weldon School of Biomedical Engineering\\
Purdue University\\
West Lafayette, USA\\
Email: euge@purdue.edu}}


\maketitle

\begin{abstract}
Pixel-wise semantic segmentation for visual scene understanding not only needs to be accurate, but also efficient in order to find any use in real-time application.
Existing algorithms even though are accurate but they do not focus on utilizing the parameters of neural network efficiently.
As a result they are huge in terms of parameters and number of operations; hence slow too.
In this paper, we propose a novel deep neural network architecture which allows it to learn without any significant increase in number of parameters.
Our network uses only 11.5 million parameters and 21.2 GFLOPs for processing an image of resolution $3\times640\times360$.
It gives state-of-the-art performance on CamVid and comparable results on Cityscapes dataset.
We also compare our networks processing time on NVIDIA GPU and embedded system device with existing state-of-the-art architectures for different image resolutions.
\end{abstract}

%
\IEEEpeerreviewmaketitle


\section{Introduction}

Recent advancement in machines with ability to perform computationally intensive tasks have enabled researchers to tap deeper into neural networks.
Convolutional neural networks' (CNNs) \cite{lecun98cnn,lecun98bp} recent success has been demonstrated in
image classification \cite{ranzato07,alex12,simonyan14,he15resnet,christian15,szegedy16inception}, localization \cite{sermanet13,tompson15}, scene understanding \cite{sturgess09,eigen15} etc.
A lot of researchers have shifted their focus towards scene understanding because of the surge in tasks like augmented reality and self-driving vehicle,
and one of the main step involved in it is pixel-level classification/semantic segmentation \cite{ren12,clement13}.

Inspired by auto-encoders \cite{ranzato07,ngiam11}, most of the existing techniques for semantic segmentation use encoder-decoder pair as core of their network architecture.
Here the encoder encodes information into feature space, and the decoder maps this information into spatial categorization to perform segmentation.
Even though semantic segmentation targets application that require real-time operation, ironically most of the current deep networks require excessively large processing time.
Networks such as YOLO \cite{redmon16}, Fast RCNN \cite{ren15}, SSD \cite{liu16} focus on real-time object detection but there is very little to no work done in this direction in case of semantic segmentation \cite{paszke16}.

In our work, we have made an attempt to get accurate instance level prediction without compromising processing time of the network.
Generally, spatial information is lost in the encoder due to pooling or strided convolution is recovered by using the pooling indices or by full convolution.
We hypothesize and later prove in our paper that instead of the above techniques; bypassing spatial information, directly from the encoder to the corresponding decoder improves accuracy along with significant decrease in processing time.
In this way, information which would have been otherwise lost at each level of encoder is preserved, and no additional parameters and operations are wasted in relearning this lost information.

In Section \ref{architecture} we give detailed explanation of our LinkNet architecture.
The proposed network was tested on popular datasets: Cityscapes \cite{cityscapes16} and CamVid \cite{camvid08} and its processing times was recorded on NVIDIA Jetson TX1 Embedded Systems module as well as on Titan X GPU.
These results are reported in Section \ref{results}, which is followed by conclusion.

\section{Related work}

Semantic segmentation involves labeling each and every pixel of an image and therefore, retaining spatial information becomes utmost important.
A neural network architecture used for scene parsing can be subdivided into encoder and decoder networks, which are basically discriminative and generative networks respectively.
State-of-the-art segmentation networks, generally use categorization models which are mostly winners of ImageNet Large Scale Visual Recognition Challenge (ILSCRC) as their discriminator.
The generator either uses the stored pooling indices from discriminator, or learns the parameters using convolution to perform upsampling.
Moreover, encoder and decoder can be either symmetric (same number of layers in encoder and decoder with same number of pooling and unpooling layers), or they can be asymmetric.

In \cite{badrinarayanan15basic} a pre-trained VGG was used as discriminator.
Pooling indices after every max-pooling step was saved and then later used for upsampling in the decoder.
Later on researchers came up with the idea of deep deconvolution network \cite{noh15learning,badrinarayanan15}, fully convolutional network (FCN) combined with skip architecture \cite{long15}, which eliminated the need of saving pooling indices.
Networks designed for classification and categorization mostly use fully connected layer as their classifier; in FCN they get replaced with convolutional layers.
Standard pre-trained encoders such as: AlexNet \cite{alex12}, VGG \cite{simonyan14}, and GoogLeNet \cite{christian15} have been used for segmentation.
In order to get precise segmentation boundaries, researchers have also tried to cascade their deep convolutional neural network (DCNN) with post-processing steps, like the use of Conditional Random Field (CRF) \cite{liang14,sturgess09}.

Instead of using networks which were designed to perform image classification, \cite{yu15dilated} proposed to use networks specifically designed for dense predictions.
Most of these networks failed to perform segmentation in real-time on existing embedded hardware.
Apart from this, recently recurrent neural networks (RNNs) were used to get contextual information \cite{visin16} and to optimize CRF \cite{zheng15}; but the use of RNN in itself makes it computationally very expensive.
Some work was also done in designing efficient network \cite{song15,paszke16}, where DCNN was optimized to get a faster forward processing time but with a decrease in prediction accuracy.

\section{Network architecture} \label{architecture}

\begin{figure}[!t]
  \centering
  \includegraphics[width=0.45\textwidth,]{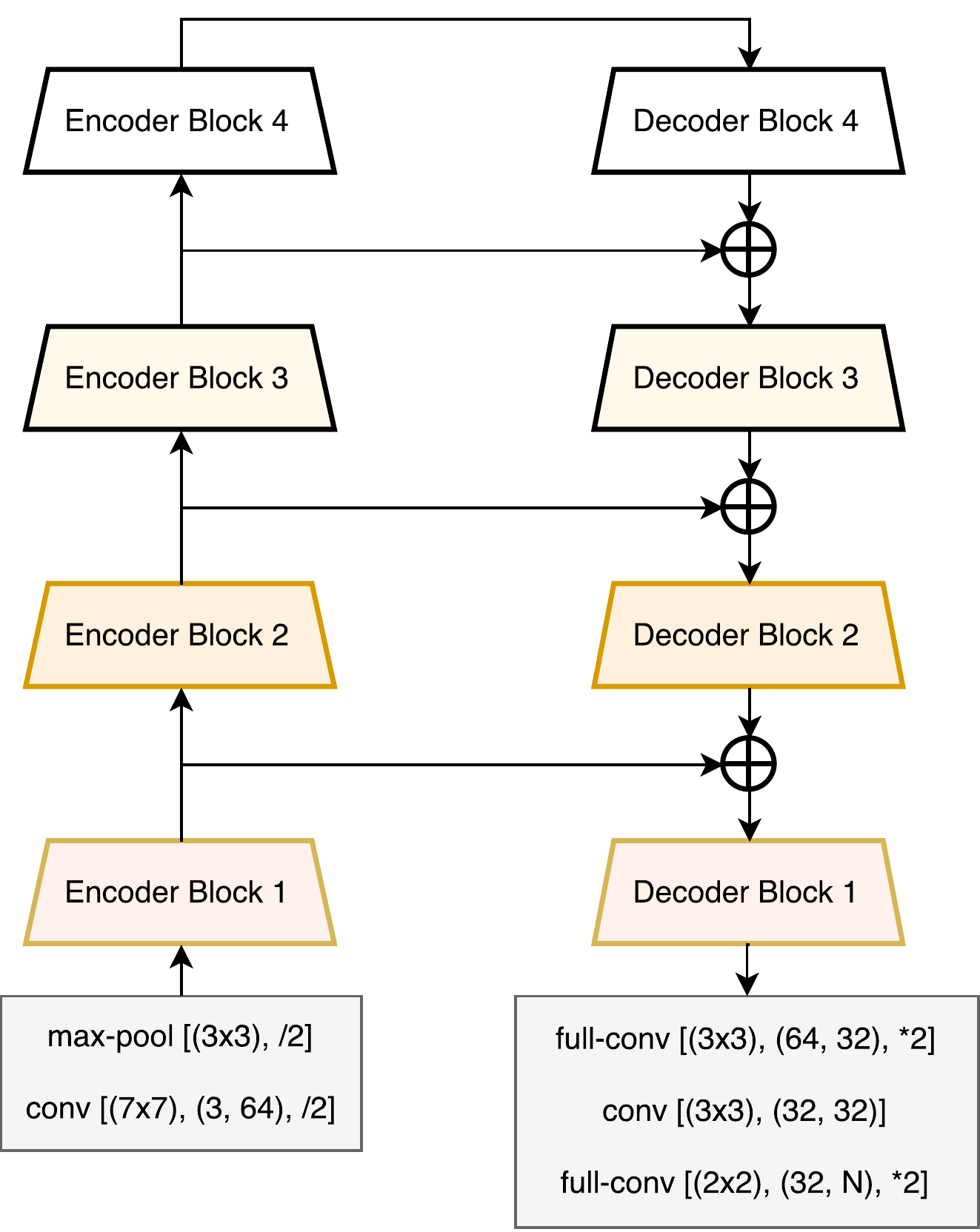}
  \vspace{0.1in}
  \caption{LinkNet Architecture}
  \vspace{-0.1in}
  \label{fig:netArch}
\end{figure}

\begin{figure}[!t]
  \centering
  \includegraphics[width=0.25\textwidth]{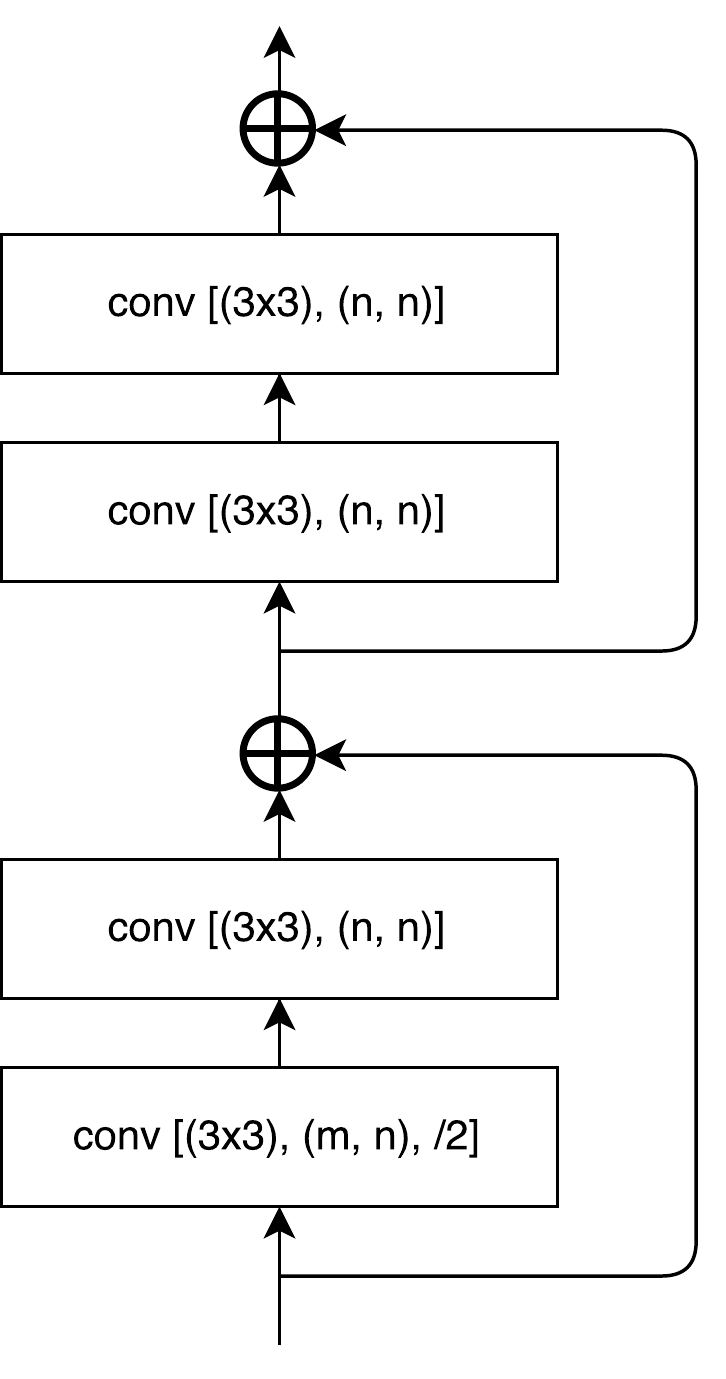}
  \caption{Convolutional modules in \textit{encoder-block (i)}}
  \label{fig:resBlock}
\end{figure}

\begin{figure}[t]
  \centering
  \includegraphics[width=0.22\textwidth]{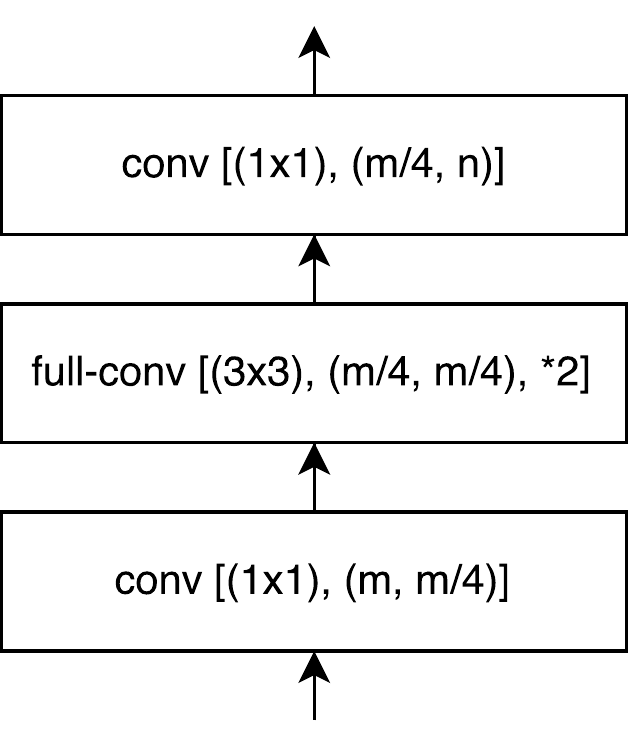}
  \caption{Convolutional modules in \textit{decoder-block (i)}}
  \label{fig:decoder}
\end{figure}

\begin{table}[!b]
  \footnotesize
  \caption{Input and output feature maps}
  \label{tab:maps}
  \centering
  \begin{tabular}{ r r r r r r r }
    \toprule
    \multirow{2}{*}{Block} & &\multicolumn{2}{c}{Encoder} &  &\multicolumn{2}{c}{Decoder} \\
    \cmidrule{3-4} \cmidrule{6-7}
    && m & n & & m & n \\
    \midrule
    1.  &&64    &64     & &64     &64   \\
    2.  &&64    &128    & &128    &64   \\
    3.  &&128   &256    & &256    &128  \\
    4.  &&256   &512    & &512    &256  \\
    \bottomrule
  \end{tabular}
\end{table}

\newcommand{\resolution}[2]{\multicolumn{2}{c}{#1$\times$#2}}
\newcommand{\ms}{\multicolumn{1}{c}{ms}}
\newcommand{\fps}{\multicolumn{1}{c}{fps}}
\begin{table*}[!t]
  \caption{Performance comparison. Image size is W$\times$H}
  \label{tab:speed}
  \centering
  \begin{tabular}{ l r r r r r r r r r r r r r r }
    \toprule
    \multirow{3}{*}{Model} &&\multicolumn{6}{c}{NVIDIA TX1} &  &\multicolumn{6}{c}{NVIDIA Titan X} \\
    \cmidrule{3-8} \cmidrule{10-15}
            &&\resolution{480}{320} &\resolution{640}{360}&\resolution{1280}{720} &
            &\resolution{640}{360}&\resolution{1280}{720}&\resolution{1920}{1080} \\
            &&\ms &\fps &\ms &\fps &\ms &\fps & &\ms&\fps &\ms&\fps &\ms&\fps \\
    \midrule
    SegNet       &&757    &1.3     &1251   &0.8   &-    &-    & &69     &14.6   &289  &3.5    &637    &1.6    \\
    ENet         &&47     &21.1    &69     &14.6  &262  &3.8  & &7      &135.4  &21   &46.8   &46     &21.6   \\
    LinkNet      &&108    &9.3     &134    &7.8   &501  &2.0  & &15     &65.8   &53   &18.7   &117    &8.5    \\
    \bottomrule
  \end{tabular}
\end{table*}

\begin{table}[htb]
  \caption{Comparison on the basis of operations}
  \label{tab:ops}
  \centering
  \begin{tabular}{ l r r r }
    \toprule
    &\multicolumn{1}{c}{GFLOPs} &\multicolumn{1}{c}{Parameters} & \multicolumn{1}{c}{Model size (fp16)}\\
    \midrule
    SegNet          &286.0                     &29.5M              &56.2 MB\\
    ENet            &3.8                       &0.4M               &0.7 MB\\
    Proposed Net    &21.2                      &11.5M              &22.0 MB\\
    \bottomrule
  \end{tabular}
\end{table}

The architecture of LinkNet is presented in Figure \ref{fig:netArch}.
Here, \texttt{conv} means convolution and \texttt{full-conv} means full convolution \cite{long15}.
Furthermore, $/2$ denotes downsampling by a factor of 2 which is achieved by performing strided convolution, and $*2$ means upsampling by a factor of 2.
We use batch normalization between each convolutional layer and which is followed by ReLU non-linearity \cite{nair10,ioffe15batchnorm}.
Left half of the network shown in Figure \ref{fig:netArch} is the encoder while the one the right is the decoder.
The encoder starts with an initial block which performs convolution on input image with a kernel of size $7\times7$ and a stride of $2$.
This block also performs spatial max-pooling in an area of $3 \times 3$ with a stride of 2.
The later portion of encoder consists of residual blocks \cite{he15resnet} and are represented as \textit{encoder-block(i)}.
Layers within these \textit{encoder-block}s are shown in detail in Figure \ref{fig:resBlock}.
Similarly, layer details for \textit{decoder-block}s are provided in Figure \ref{fig:decoder}.
Table \ref{tab:maps} contains the information about the feature maps used in each of these blocks.
Contemporary segmentation algorithms use networks such as VGG16 (138 million parameters), ResNet101 (45 million parameters) as their encoder which are huge in terms of parameters and GFLOPs.
LinkNet uses ResNet18 as its encoder, which is fairly lighter network and still outperforms them as evident from Section \ref{results}.
We use the technique of full-convolution in our decoder as proposed earlier by \cite{long15}.
Every \texttt{conv}$(k\times k)(im,om)$ and \texttt{full-conv}$(k\times k)(im,om)$ operations has at least three parameters.
Here, $(k\times k)$ represent $(kernel-size)$ and $(im,om)$ represent $(input map, output map)$ respectively.

Unlike existing neural network architectures which are being used for segmentation, our novelty lies in the way we link each encoder with decoder.
By performing multiple downsampling operations in the encoder, some spatial information is lost.
It is difficult to recover this lost information by using only the downsampled output of encoder.
\cite{badrinarayanan15basic} linked encoder with decoder through pooling indices, which are not trainable parameters.
Other methods directly use the output of their encoder and feed it into the decoder to perform segmentation.
In this paper, input of each encoder layer is also bypassed to the output of its corresponding decoder.
By doing this we aim at recovering lost spatial information that can be used by the decoder and its upsampling operations.
In addition, since the decoder is sharing knowledge learnt by the encoder at every layer, the decoder can use fewer parameters.
This results in an overall more efficient network when compared to the existing state-of-the-art segmentation networks, and thus real-time operation.
Information about trainable parameters and number operations required for each forward pass is provided in detail in Section \ref{results}.

\section{Results} \label{results}

We compare LinkNet with existing architectures on two different metrics:
\begin{enumerate}
    \item Performance in terms of speed:
        \begin{itemize}
            \item Number of operations required to perform one forward pass of the network
            \item Time taken to perform one forward pass
        \end{itemize}
    \item Performace interms of accuracy on Cityscapes \cite{cityscapes16} and CamVid \cite{camvid08} datasets.
\end{enumerate}

\subsection{Performance Analysis}

We report inference speed of LinkNet on NVIDIA TX1 embedded system module as well as on widely used NVIDIA TitanX.
Table \ref{tab:speed} compares inference time for a single input frame with varying resolution.
As evident from the numbers provided, LinkNet can process very high resolution image at 8.5 fps on GPU.
More importantly, it can give real-time performance even on NVIDIA TX1.
'-' indicates that network was not able to process image at that resolution on the embedded device.

We choose $640 \times 360$ as our default image resolution and report number of operations required to process image of this resolution in Table \ref{tab:ops}.
Number of operations determine the forward pass time of any network, therefore reduction in it is more vital than reduction in number of parameters.
Our approach's efficiency is evident in the much low number of operations per frame and overall parameters.

\subsection{Benchmarks}

We use Torch7 \cite{collobert11} machine-learning tool for training with RMSProp as the optimization algorithm.
The network was trained using four NVIDIA TitanX.
Since the classes present in all the datsets are highly imbalanced; we use a custom class weighing scheme defined as $w_{\mathrm{class}} = \frac{1}{\ln(1.02 + p_{\mathrm{class}})}$.
This class weighing scheme has been taken from \cite{paszke16} and it gave us better results than mean average frequency.
As suggested in Cityscapes \cite{cityscapes16}, we use intersections over union (IoU) and instance-level intersection over union (iIoU) as our performance metric instead of using pixel-wise accuracy.
In order to prove that the bypass connections do help, each table contains IoU and iIoU values with as well as without bypass.
We also compare LinkNet's performance with other standard models such as SegNet \cite{badrinarayanan15}, ENet \cite{paszke16}, Dilation8/10 \cite{yu15}, and Deep-Lab CRF \cite{chen16}.

\begin{table}[!b]
  \small
  \caption{Cityscapes val set results (* on test set)}
  \label{tab:cityscape}
  \centering
  \begin{tabular}{ l r r r r }
    \toprule
    Model                    &Class IoU      &Class iIoU     \\
    \midrule
    SegNet*                  &56.1           &34.2           \\
    ENet*                    &58.3           &34.4           \\
    Dilation10               &68.7           &-              \\
    Deep-Lab CRF (VGG16)     &65.9           &-              \\
    Deep-Lab CRF (ResNet101) &71.4           &42.6           \\
    LinkNet without bypass   &72.6           &51.4           \\
    LinkNet                  &\textbf{76.4}  &\textbf{58.6}  \\
    \bottomrule
  \end{tabular}
\end{table}

\begin{table*}[!t]
  \footnotesize
  \caption{Results on CamVid test set of (1) SegNet, (2) ENet, (3) Dilation8, (4) LinkNet without bypass, and (5) LinkNet}
  \label{tab:camvid}
  \centering
  \begin{tabular}{ c c c c c c c c c c c c c c }
    \toprule
    \rotatebox[origin=c]{90}{Model}\hspace{0.07in} &\rotatebox[origin=c]{90}{Building} &\rotatebox[origin=c]{90}{Tree} &\rotatebox[origin=c]{90}{Sky} &\rotatebox[origin=c]{90}{Car} &\rotatebox[origin=c]{90}{Sign} &\rotatebox[origin=c]{90}{Road} &\rotatebox[origin=c]{90}{ Pedestrian } &\rotatebox[origin=c]{90}{Fence} &\rotatebox[origin=c]{90}{Pole} &\rotatebox[origin=c]{90}{Sidewalk} &\rotatebox[origin=c]{90}{Bicyclist} &\hspace{0.07in}\rotatebox[origin=c]{90}{IoU} &\rotatebox[origin=c]{90}{iIoU} \\
    \midrule
    1\hspace{0.07in}    &88.8   &87.3   &92.4   &82.1   &20.5   &97.2   &57.1   &49.3   &27.5   &84.4   &30.7   &\hspace{0.07in}65.2             &55.6      \\
    3\hspace{0.07in}    &74.7   &77.8   &95.1   &82.4   &51.0   &95.1   &67.2   &51.7   &35.4   &86.7   &34.1   &\hspace{0.07in}68.3             &51.3      \\
    2\hspace{0.07in}    &82.6   &76.2   &89.9   &84.0   &46.9   &92.2   &56.3   &35.8   &23.4   &75.3   &55.5   &\hspace{0.07in}65.3             &-         \\
    3\hspace{0.07in}    &84.6   &87.4   &88.8   &72.6   &37.1   &95.3   &61.2   &56.0   &33.1   &88.3   &24.4   &\hspace{0.07in}66.3             &52.7      \\
    4\hspace{0.07in}    &88.8   &85.3   &92.8   &77.6   &41.7   &96.8   &57.0   &57.8   &37.8   &88.4   &27.2   &\hspace{0.07in}\textbf{68.3}    &\textbf{55.8} \\
    \bottomrule
  \end{tabular}
\end{table*}

\begin{table*}[!t]
  \begin{adjustbox}{width=0.90\textwidth,center}
    \begin{tabular}{MMM}
       \includegraphics[width=.20\textwidth]{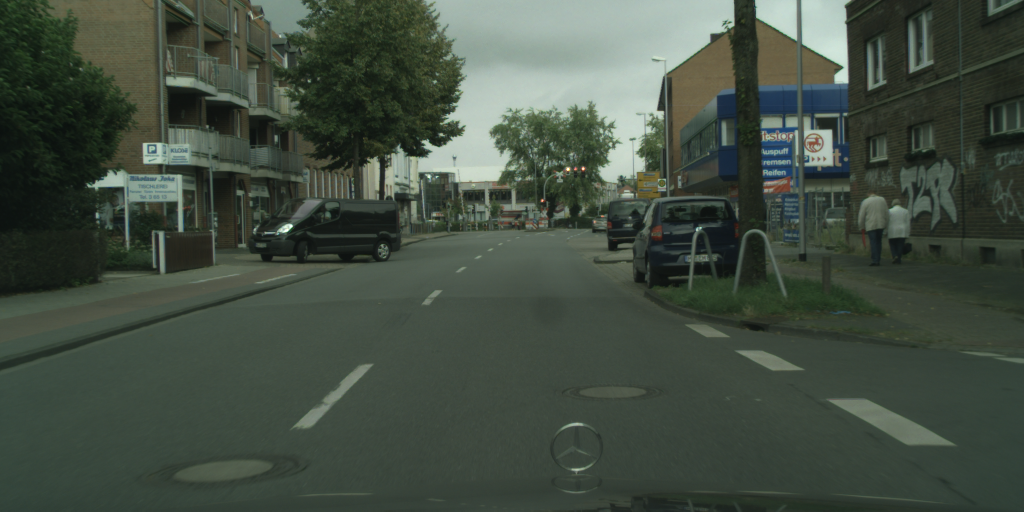}
      &\includegraphics[width=.20\textwidth]{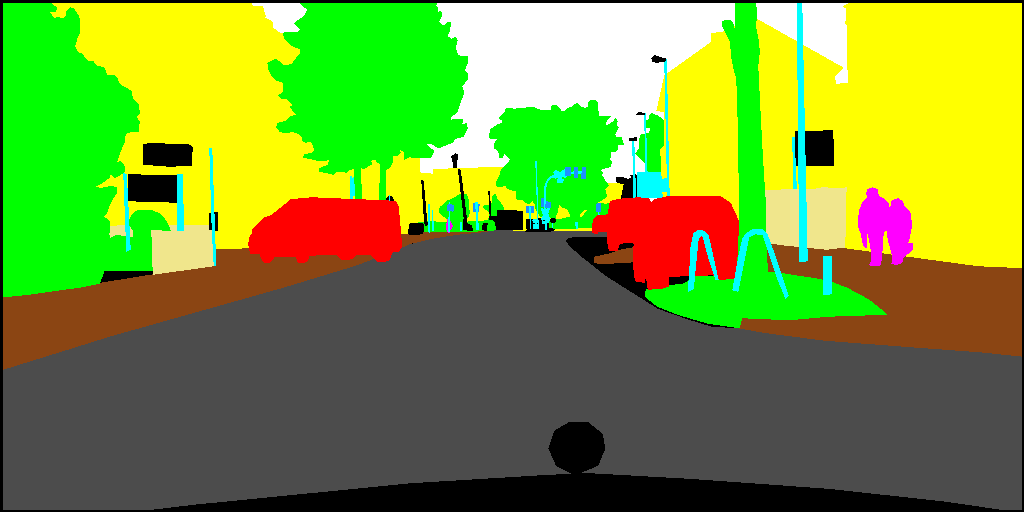}
      &\includegraphics[width=.20\textwidth]{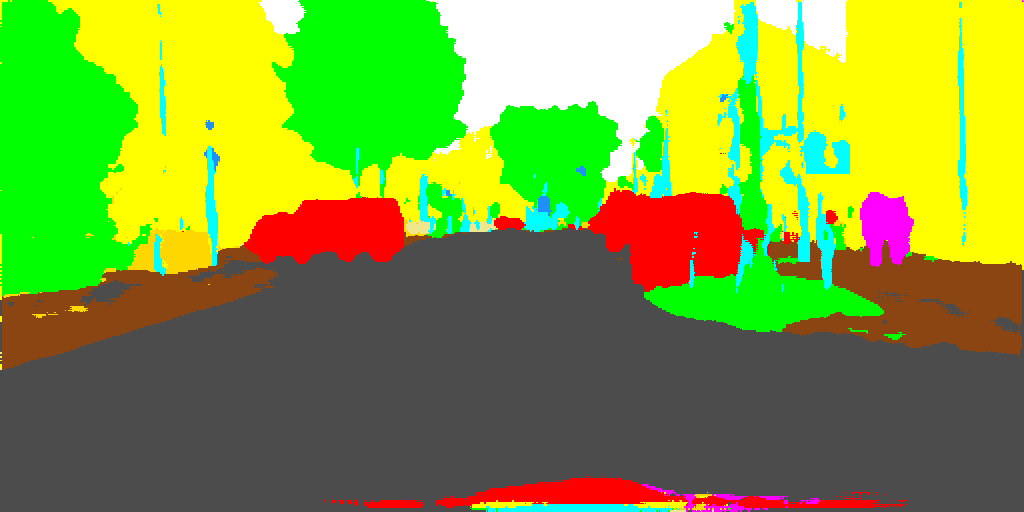}
      \\
       \includegraphics[width=.20\textwidth]{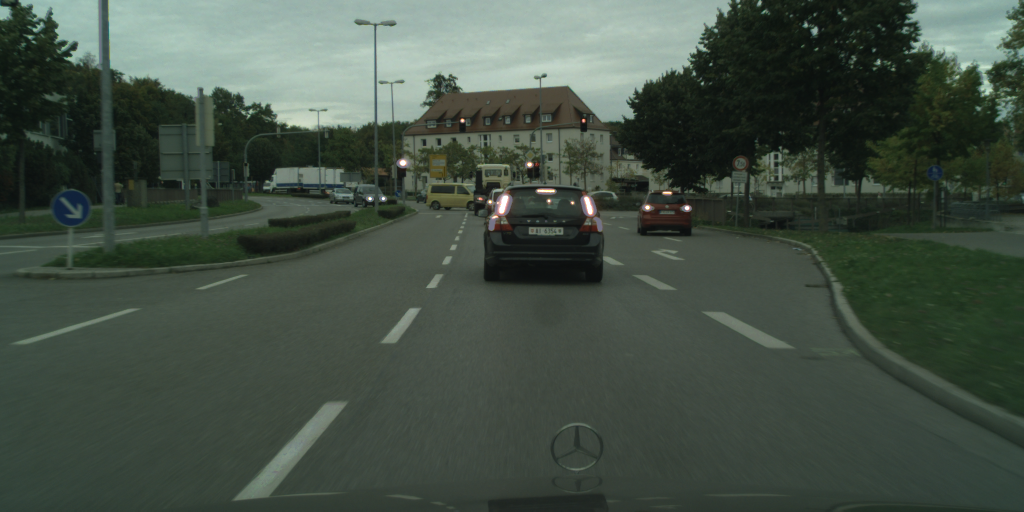}
      &\includegraphics[width=.20\textwidth]{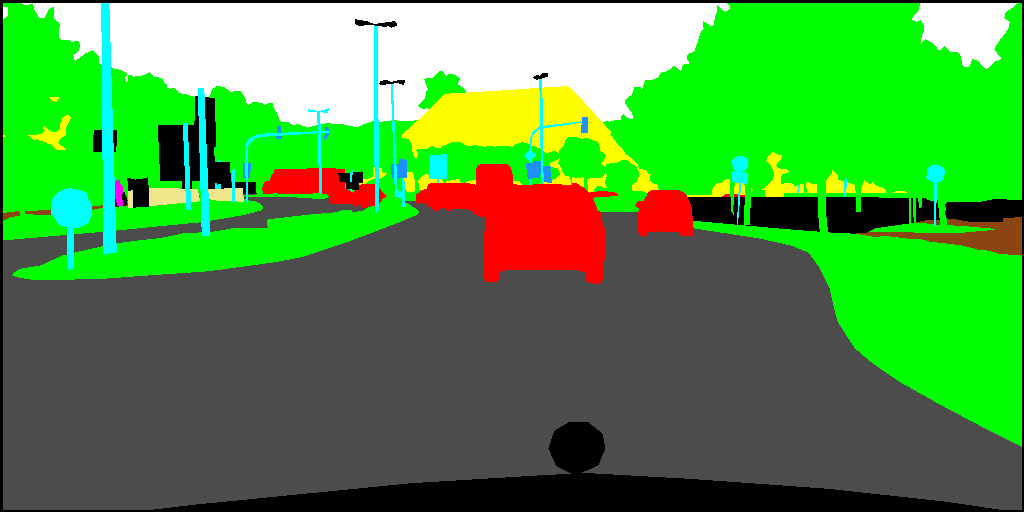}
      &\includegraphics[width=.20\textwidth]{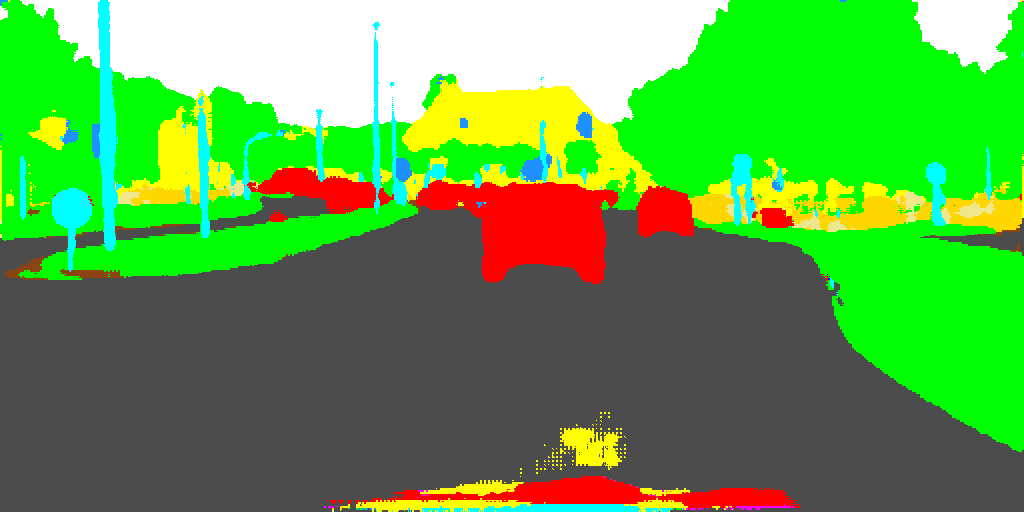}
      \\
      \scriptsize(a) Input Image &\scriptsize(b) Ground truth &\scriptsize(c) Prediction
    \end{tabular}
  \end{adjustbox}
  \captionof{figure}{LinkNet prediction on Cityscapes \cite{cityscapes16} test set.}
  \label{fig:resultcs}
\end{table*}

\begin{table*}[!t]
  \begin{adjustbox}{width=0.90\textwidth,center}
    \begin{tabular}{MMM}
       \includegraphics[width=.20\textwidth]{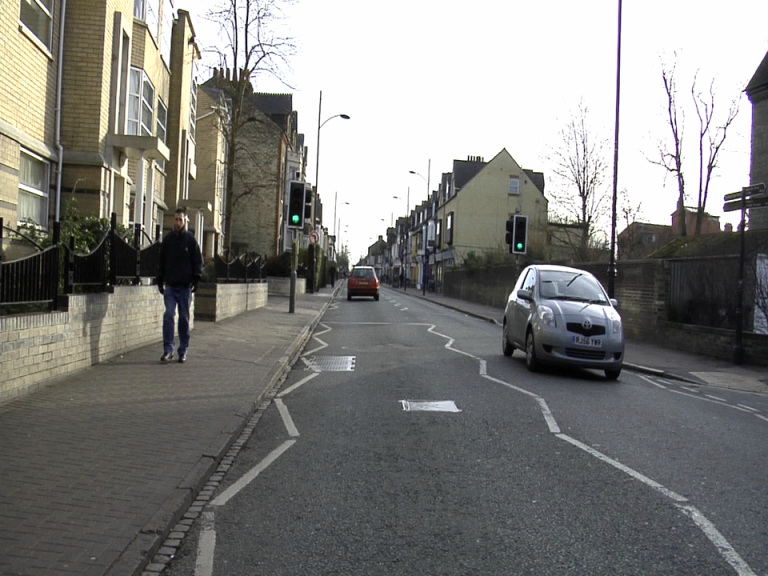}
      &\includegraphics[width=.20\textwidth]{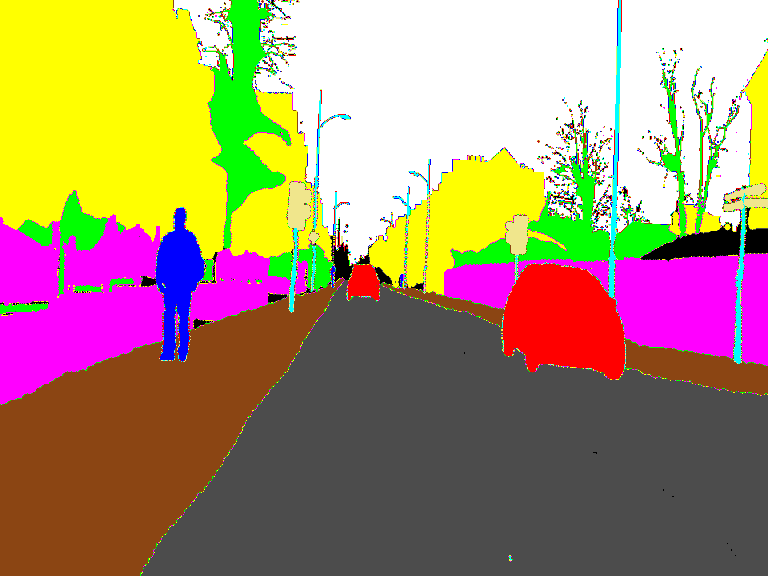}
      &\includegraphics[width=.20\textwidth]{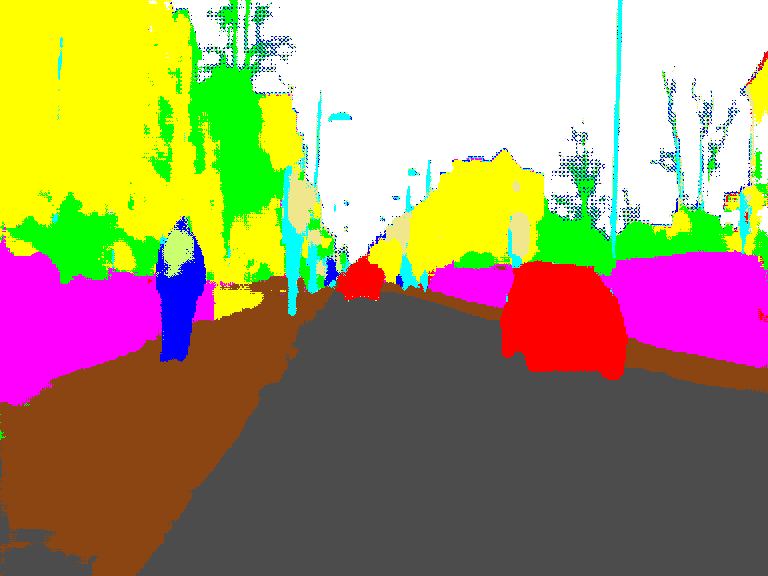}
      \\
       \includegraphics[width=.20\textwidth]{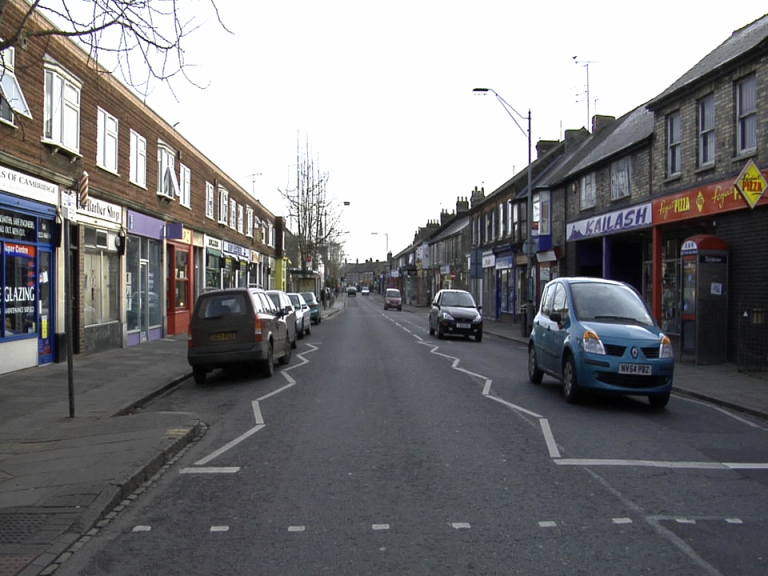}
      &\includegraphics[width=.20\textwidth]{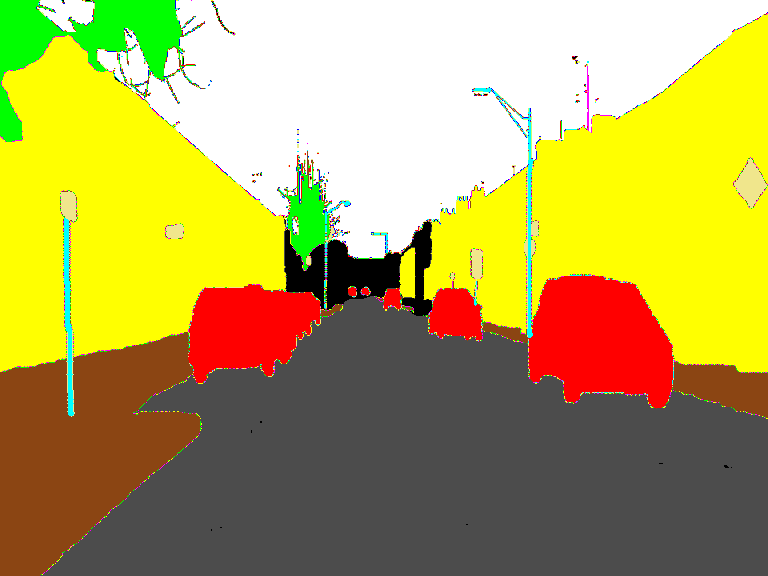}
      &\includegraphics[width=.20\textwidth]{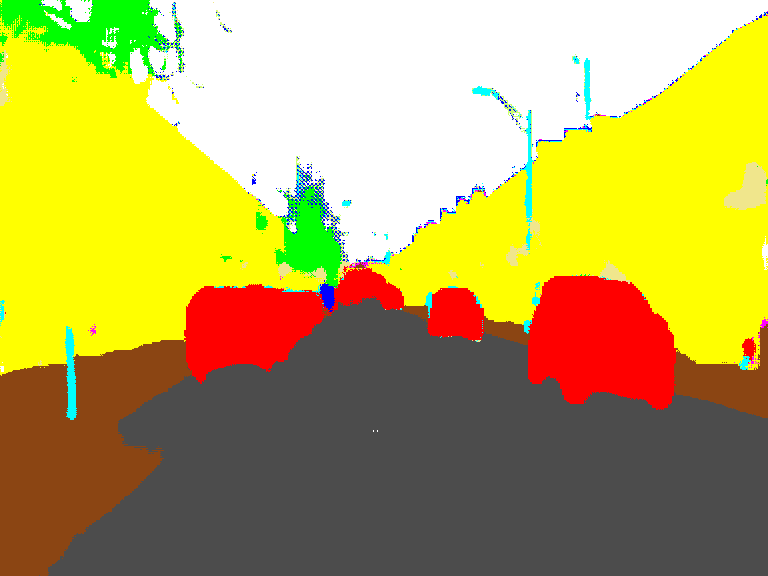}
      \\
      \scriptsize(a) Input Image &\scriptsize(b) Ground truth &\scriptsize(c) Prediction
    \end{tabular}
  \end{adjustbox}
  \captionof{figure}{LinkNet prediction on CamVid \cite{camvid08} test set.}
  \label{fig:resultcv}
\end{table*}

\paragraph{Cityscapes}
This dataset consists of 5000 fine-annotated images, out of which 2975 are available for training, 500 for validation, and the remaining 1525 have been selected as test set \cite{cityscapes16}.
We trained on our network on 19 classes that was provided in the official evaluation scripts \cite{cityscapes16}.
As reported in Table \ref{tab:cityscape}, our network outperforms existing models.
These performance values were calculated on validation dataset.
Input image of resolution $1024 \times 512$ was used for training the network.
A batch size of 10 and initial learning rate of $5\mathrm{e}{-4}$ was found to give optimum performance.
Figure \ref{fig:resultcs} shows the predicted segmented output on couple of cityscapes test images.

\paragraph{CamVid}

It is another automotive dataset which contains 367 training, 101 validation, and 233 testing images \cite{camvid08}.
There are eleven different classes such as building, tree, sky, car, road, etc. while the twelfth class contains unlabeled data, which we ignore during training.
The original frame resolution for this dataset is $960 \times 720$ (W,H) but we downsampled the images by a factor of 1.25 before training.
Due to hardware constraint, batch size of 8 was used to train the network.
In Table \ref{tab:camvid} we compare the performance of the proposed algorithm with existing state-of-the-art algorithms on test set.
LinkNet outperforms all of them in both IoU and iIoU metrics.
Segmented output of LinkNet can be seen in Figure \ref{fig:resultcv}

\section{Conclusion} \label{conclusion}

We have proposed a novel neural network architecture designed from the ground up specifically for semantic segmentation.
Our main aim is to make efficient use of scarce resources available on embedded platforms, compared to fully fledged deep learning workstations.
Our work provides large gains in this task, while matching and at times exceeding existing baseline models, that have an order of magnitude larger computational and memory requirements.
The application of proposed network on the NVIDIA TX1 hardware exemplifies real-time portable embedded solutions.

Even though the main goal was to run the network on mobile devices, we have found that it is also very efficient on high end GPUs like NVIDIA Titan X.
This may prove useful in data-center applications, where there is a need of processing large numbers of high resolution images.
Our network allows to perform large-scale computations in a much faster and more efficient manner, which might lead to significant savings.

\section*{Acknowledgment}

This work was partly supported by the Office of Naval Research (ONR) grants N00014-12-1-0167, N00014-15-1-2791 and MURI N00014-10-1-0278.
We gratefully acknowledge the support of NVIDIA Corporation with the donation of the TX1, Titan X, K40 GPUs used for this research.

\bibliographystyle{IEEEtran}
{
\fontsize{9}{9}
\selectfont
\bibliography{ref}
}

\end{document}